\documentclass[runningheads]{llncs}

\usepackage{times}
\usepackage{epsfig}
\usepackage{graphicx}
\usepackage{adjustbox}
\usepackage{amsmath}
\usepackage{amssymb}
\usepackage{booktabs}
\usepackage{multirow}
\usepackage[table,xcdraw]{xcolor}
\usepackage{dsfont}
\newcommand{\cmark}{\ding{51}} %
\newcommand{\xmark}{\ding{55}} %
\usepackage{pifont}
\usepackage[T1]{fontenc}
\usepackage{graphicx}
\usepackage{hyperref}
\usepackage{color}

\begin{document}
\title{FastTextSpotter: A High-Efficiency Transformer for Multilingual Scene Text Spotting}
%
%


\author{Alloy Das\textsuperscript{\textsection}\inst{1} \and
Sanket Biswas\textsuperscript{\textsection}\inst{2} \and
Umapada Pal\inst{1}\and
Josep Llad\'{o}s\inst{2}\and
Saumik Bhattacharya\inst{3}
}
\authorrunning{A. Das et al.}
%
\institute{CVPR Unit, Indian Statistical Institute, Kolkata\\ 
\email{alloyuit@gmail.com, umapada@isical.ac.in}\\ \and
Computer Vision Center, Universitat Autònoma de Barcelona\\
\email{\{sbiswas,josep\}@cvc.uab.cat }\and
ECE, Indian Institute of Technology, Kharagpur\\
\email{saumik@ece.iitkgp.ac.in}}

\maketitle              
    \begingroup\renewcommand\thefootnote{\textsection}
\footnotetext{Equal contribution}
\endgroup

\begin{abstract}
    The proliferation of scene text in both structured and unstructured environments presents significant challenges in optical character recognition (OCR), necessitating more efficient and robust text spotting solutions. This paper presents FastTextSpotter, a framework that integrates a Swin Transformer visual backbone with a Transformer Encoder-Decoder architecture, enhanced by a novel, faster self-attention unit, SAC2, to improve processing speeds while maintaining accuracy. FastTextSpotter has been validated across multiple datasets, including ICDAR2015 for regular texts and CTW1500 and TotalText for arbitrary-shaped texts, benchmarking against current state-of-the-art models. Our results indicate that FastTextSpotter not only achieves superior accuracy in detecting and recognizing multilingual scene text (English and Vietnamese) but also improves model efficiency, thereby setting new benchmarks in the field. This study underscores the potential of advanced transformer architectures in improving the adaptability and speed of text spotting applications in diverse real-world settings. \textcolor{black}{The dataset, code, and pre-trained models have been released in our \href{https://github.com/alloydas/Fast-Textspotter}{\texttt{Github}}}.


\end{abstract}


%
\section{Introduction}

In the rapidly evolving field of pattern recognition, text spotting— the task of localizing and recognizing text within natural scenes — poses unique challenges. These challenges have been addressed through powerful optical character recognition (OCR) systems designed to handle text in both structured~\cite{garcia2024step} and unstructured~\cite{atienza2021vision,fang2022abinet++} environments. These environments commonly feature text in multiple ranges of orientations (from arbitrary-shaped~\cite{ch2020total,liu2019curved,singh2021textocr} to regular-shaped~\cite{karatzas2015icdar}), annotation styles (from rotated quadrilaterals~\cite{karatzas2015icdar}, polygonal word-level~\cite{ch2020total}, polygonal sentence-level~\cite{liu2019curved} to hierarchical layout-level~\cite{long2022towards}) and diverse language domains (multilingual~\cite{nayef2017icdar2017,nayef2019icdar2019} to low-resource languages~\cite{nguyen2021dictionary} to different scripts~\cite{zhang2019icdar}). The overall computational load of processing such high-resolution images to detect and recognize text accurately across different text orientations, languages and styles is substantial.

 \begin{figure}[t]
    \centering
    \includegraphics[width = 0.48\textwidth, height= 5cm]{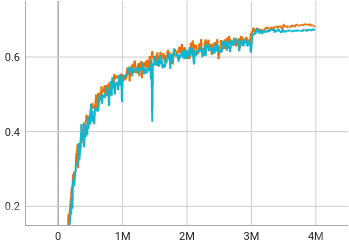}
    \caption{\textbf{Trade-off between text spotting performance h-mean vs number of training iterations}: The blue curve indicates the model without the SAC2 attention module while the orange curve depicts the model performance with our proposed SAC2 module.}
    \label{fig:fig1}
\end{figure}

Current state-of-the-art models have made significant contributions towards improving text detection and recognition capabilities, which employs both Convolutional Neural Networks (CNNs)~\cite{liu2020abcnet,liu2021abcnet,qiao2021mango,qiao2020text} and Transformers~\cite{huang2022swintextspotter,zhang2022text,fang2022abinet++,das2024harnessing,ye2023deepsolo}. Despite significant advancements, these models typically struggle with balancing high accuracy and computational efficiency, especially under constrained resources or in real-time applications. This is particularly critical in scenarios where quick text interpretation is essential, such as in navigation aids for visually impaired individuals or instant translation services. Fast and reliable text interpretation can indeed help bridge the digital divide by making information more accessible to people who speak less common languages or dialects. Moreover, it can automate and improve the process of annotating vast amounts of data, which is essential for training more robust OCR systems that provide higher-quality, context-aware annotations. 

The emergence of text spotting transformers (TESTR)~\cite{zhang2022text} has prompted the adoption of detection transformer (DETR) architectures~\cite{carion2020end} as foundational backbones within text spotting frameworks as in~\cite{das2024harnessing,ye2023deepsolo}. Recent methods~\cite{ye2023deepsolo,ye2023dptext} have focused on enabling more efficient training and faster convergence using deformable attention~\cite{zhu2020deformable} on the dynamic control point queries of text coordinates. Using point coordinates to obtain positional queries rather than anchor boxes, as described in~\cite{zhang2022text}, allows the transformer decoder to dynamically update points for scene text detection. In this work, we explore the possibility of extending this dynamic attention mechanism towards the task of scene text spotting. Moreover, recent works~\cite{das2024harnessing,huang2022swintextspotter,das2024diving} have recently shown the effectiveness of Swin Transformers~\cite{liu2021swin} by generating hierarchical feature maps which are critical for fine-grained predictions necessary in text segmentation. This work introduces \textbf{FastTextSpotter}, a novel text spotting framework that combines a Swin-tiny backbone for visual feature extraction with a dual-decoder transformer encoder architecture~\cite{zhang2022text} tailored for text spotting. To optimize training, we introduce a novel attention module, \textbf{SAC2} (Self-Attention with Circular Convolutions), inspired by~\cite{ye2023dptext,peng2020deep}. This novel component, integrated within our text spotting framework, not only competes well with existing state-of-the-art (SOTA) text spotters but also enhances operational efficiency, particularly in frames per second (FPS), setting a new benchmark in text spotting performance. Figure~\ref{fig:fig1} illustrates the accuracy vs efficiency trade-off and the impact of the SAC2 attention module. In this context, we also explore the following key research questions to gain a more comprehensive understanding of the trade-offs between accuracy and efficiency in SOTA text spotting models. 1) How can the computational efficiency of attention mechanisms in Transformer models be improved without compromising on text detection and recognition accuracy? 2) What architectural modifications are necessary to adapt the Swin Transformer for optimal performance in diverse text spotting scenarios and orientations? 3) Can the model effectively handle multilingual text spotting, particularly in low-resource languages like Vietnamese?

The paper proposes a three-fold contribution: 1) We develop FastTextSpotter, integrating a Swin-Tiny backbone with an efficient text spotting setup, significantly enhancing scene text detection and recognition efficiency.
2) We introduce SAC2, a dynamic attention mechanism that accelerates training and convergence while maintaining high detection accuracy. 
3) FastTextSpotter excels in detecting multilingual and variably oriented texts, including in resource-limited languages like Vietnamese, broadening its utility in diverse applications.

\section{Related Work}

Our FastTextSpotter framework is designed to refer to the previous works introduced below, aiming to handle scene text spotting in an efficient way that combines text spotting transformers with a dynamic and faster attention module.

\noindent
\textbf{End-to-End Object Detection.} 
 The DETR approach~\cite{carion2020end} proposed the first end-to-end transformer-based object detection as a set prediction task without complex hand-crafted anchor generation and post-processing. The Deformable-DETR~\cite{zhu2020deformable} addressed the task by attending to sparse features using a deformable cross-attention operation which reduces the quadratic complexity of DETR to linear complexity and leveraging multi-scaled features. DE-DETR~\cite{wang2022towards} identifies that the main element which impacts the model efficiency is related to the sparse feature sampling, whereas DAB-DETR~\cite{liu2022dabdetr} handled the above issue using dynamic anchor boxes as position queries. In our study, we recast this query in point formulation to modify the Transformer Decoder backbone in TESTR~\cite{zhang2022text} for both detection and recognition tasks to handle arbitrarily shaped scene texts and also fasten the training process. Recent methods like ~\cite{zheng2023less} focus attention on more information-rich tokens for improving trade-off between efficiency and model performance.\\  

\noindent
\textbf{Scene Text Spotting.} 
The rise of deep learning has significantly advanced the field of scene text spotting. Early methods treated text spotting as a two-stage process, training separate detection and recognition modules that were combined at inference time~\cite{wang2011end,liao2017textboxes}. More recent approaches have adopted end-to-end strategies~\cite{li2017towards,liu2018fots}, simultaneously tackling detection and recognition through RoI operations to address arbitrary-shaped texts with some using quadrangle text region proposals~\cite{sun2019textnet,feng2019textdragon}. Other approaches employ the MaskTextSpotter~\cite{lyu2018mask,liao2020mask} series which employed binary maps for text and character-level segmentation tasks based on Mask-RCNN~\cite{he2017mask} to reduce segmentation errors. PAN++~\cite{wang2021pan++} have further refined these methods by enhancing segmentation efficiency and reducing background interference, similar to those adopted in ~\cite{qin2019towards,wang2019arbitrary}. While these approaches produced acceptable performance, the mask representation needed some further post-processing steps. MANGO~\cite{qiao2021mango} proposed a mask attention module to utilize global features across several text instances, however, it required center-line segmentation for the predictions. Other attempts to create customized representations for curved texts include Parametric Bezier curves ~\cite{liu2020abcnet,liu2021abcnet}, Shape Transform module ~\cite{qiao2020text} etc. \\

\noindent
\textbf{Impact of Transformers.} The introduction of Transformers~\cite{vaswani2017attention} has marked a pivotal shift towards transformer-based architectures in text spotting, eliminating the need for RoI operations by leveraging global feature modelling, as seen in applications like ABINet and SwinTextSpotter~\cite{fang2021read,huang2022swintextspotter}. 
The introduction of vision transformers~\cite{dosovitskiy2020image} also opened the floodgates to its application in STR application~\cite{atienza2021vision}.  ABINet~\cite{fang2022abinet++,fang2021read} integrate advanced techniques such as bidirectional language modelling and Feature Pyramid Networks (FPNs) to improve text detection and recognition, particularly for texts entangled with complex backgrounds or small sizes.  Our proposed framework, leveraging a Swin-Transformer~\cite{liu2021swin} with a tiny variant for computational efficiency, utilizes a multi-scale deformable attention mechanism~\cite{zhu2020deformable} used by TESTR approach~\cite{zhang2022text} to optimize feature extraction across various text sizes \textit{without necessitating post-processing for polygon vertices or Bezier control points}, thereby streamlining the text spotting process and enhancing overall system performance.



\section{Methodology}

\begin{figure*}[ht]
\begin{center}
\includegraphics[width=.9\textwidth, height = 6.5cm]{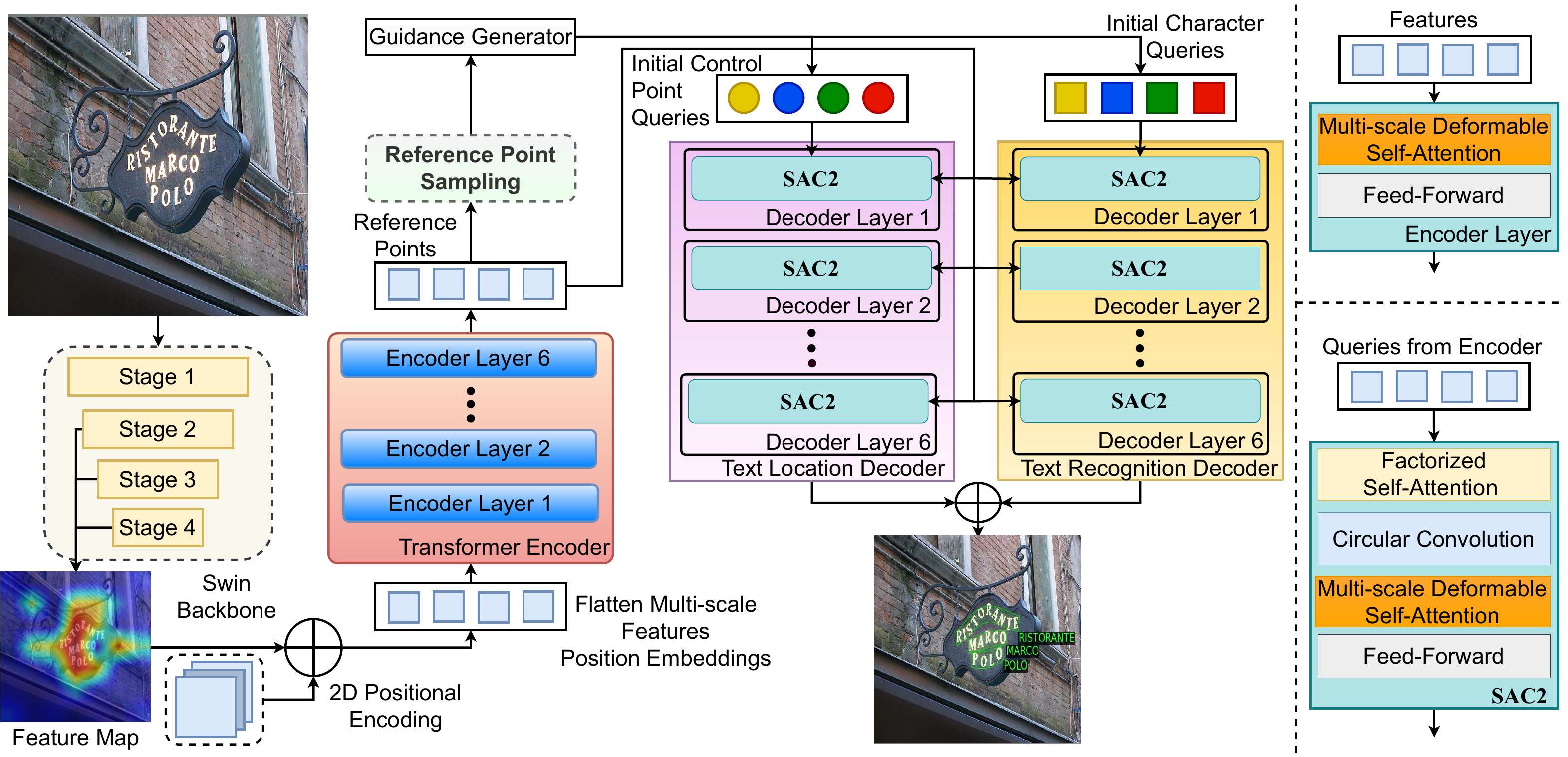}
\end{center}
   \caption{\textbf{Overview of FastTextSpotter} illustrating a Swin Transformer visual backbone with a Transformer Encoder-Decoder framework. Key features include the SAC2 attention module, dual decoders for accurate text localization and recognition, and the Reference Point Sampling system for effective text detection across various shapes and languages.}
\label{fig2}
\end{figure*}

The core objective of the FastTextSpotter framework is to enhance the efficiency and accuracy of scene text detection and recognition. This section details the architectural components and the novel self-attention unit, SAC2, which together form the backbone of our proposed system. Additionally, we outline the training objectives and processes that drive the performance of the entire framework.

\subsection{Model Architecture}
he overall architectural framework, as depicted in Fig.\ref{fig2}, is composed of three primary components: (1) a visual feature extraction unit that utilizes a Swin-Transformer\cite{liu2021swin} backbone for the extraction of multi-scale features; (2) a text spotting module that includes a Transformer encoder, which encodes the image features into positional object queries, followed by two separate Transformer decoder units that are responsible for predicting the locations of text instances and recognizing the corresponding characters.

\noindent
\textbf{Visual Feature Extraction Unit.} anilla convolutions, which operate locally at fixed sizes (e.g., $3 \times 3$), struggle to connect distant features effectively. Text spotting, however, demands the ability to capture relationships between various text regions within the same image, while also accounting for similarities in background, style, and texture. To address this, we selected a compact yet efficient Swin-Transformer~\cite{liu2021swin} unit, referred to as Swin-tiny, to extract more detailed and fine-grained image features in Fig~\ref{fig:swin_feature_map}.



\noindent
\textbf{Text Spotting Unit.} The text-spotting module primarily comprises a transformer encoder and two transformer decoders dedicated to text detection and recognition, following a schema similar to the TESTR framework~\cite{zhang2022text}. We formulate this task as a set prediction problem inspired by DETR~\cite{carion2020end}, aiming to predict a set of point-character pairs for each image. Specifically, we define it as $A = {(E^{(j)}, F^{(j)})}_{j=1}^K$, where $j$ indicates the index of each instance. Here, $E^{(j)} = {e^{(j)}_1, \dots, s^{(j}_M}$ represents the coordinates of $M$ control points, and $F^{(j)} = {f^{(j)}_1, \dots, f^{(j)}_M}$ corresponds to the sequence of $M$ text characters. In this unified framework, the text location decoder (TLD) predicts $E^{(j)}$, while the text recognition decoder (TRD) predicts $F^{(j)}$.\\

\noindent
\textbf{Text Location Decoder.} In the location decoder, queries are transformed into composite queries that predict multiple control points for each text instance. We define these as $Q$ queries, with each one corresponding to a text instance denoted as $E^{(j)}$. Each query comprises several sub-queries $e_n$, such that $e^{(j)} = {e^{(j)}_1, \dots, e^{(j)}_M}$. These initial control points are then processed through the location decoder, which consists of multiple layers. This is followed by a classification head that predicts confidence levels for the final control points, alongside a two-channel regression head that generates the normalized coordinates for each point. In this context, the control points are defined as the polygon vertices, starting from the top-left corner and proceeding in a clockwise direction.

\noindent
\textbf{Text Recognition Decoder.} The character decoder operates similarly to the location decoder, with the key difference being that control point queries are replaced with character queries, denoted as $F^{(j)}$. Both $E^{(j)}$ and $F^{(j)}$ queries, sharing the same index, correspond to the same text instance. Consequently, during the prediction phase, each decoder simultaneously predicts the control points and the characters for the corresponding instance. Finally, a classification head is employed to predict multiple character classes based on the final character queries.

\subsection{Query Point Formulation and SAC2 Attention Module}

The training efficiency of the FastTextSpotter is primarily driven by a dynamic point update strategy, which updates prediction points during sampling from the transformer encoder unit. This is followed by the application of the SAC2 attention module in the subsequent text location and recognition decoders.\\  

\noindent
\textbf{Reference Point Sampling.} 
We adopt the box-to-polygon conversion method from TESTR~\cite{zhang2022text}, which effectively transforms axis-aligned box predictions into polygon representations of scene text. This approach, inspired by ~\cite{ye2023dptext} simplifies and improves the scene text detection. Positional queries are generated from anchor boxes using a 2D positional encoding, enhanced with a multi-layered perceptron as implemented in~\cite{liu2022dabdetr}, with the objective of making these queries learnable. Specifically, these dynamic anchor boxes—post the final Transformer encoder layer—are concatenated with $M$ content queries for control points and $A$ content queries for text characters, refining the text spotting process. The following Eq.~\ref{eq:points} explains the used strategy of creating the compositional queries $Q^{(j)}(j = 1,..., K)$ : 

\begin{equation}
    Q^{(j)} =E^{(j)} + F = \theta ((s,r, c, d)^{(j)}) + (e_1, e_2, . . , e_M )
\label{eq:points}
\end{equation}

where S and R stand in for the relevant positional and content components of each composite query.
The sine positional encoding function is followed by a normalising and linear layer.
The center coordinate and scale details of each anchor box are represented by $(s, r, c, d)$.
The M learnable control point content queries shared over K composite inquiries are $(e_1,..., e_M)$. Keep in mind that we used the detector with the Eq.~\ref{eq:points} query formulation in our model.
We sample $\frac{M}{2}$ point coordinates $point_m (m = 1,..., M)$ evenly on the top and bottom side of each anchor box, respectively, motivated by the positional label form and the shape prior that the top and bottom side of a scene text are often close to the corresponding side on bounding box as in Eq.~\ref{eq:points1}: 
\begin{equation}
point_n=\begin{cases}
(s - \frac{c}{2} + \frac{(m - 1)\times c}{\frac{M}{2} - 1}, r- \frac{d}{2}), & m \leqslant \frac{M}{2} \\
(s - \frac{c}{2} + \frac{(M - m)\times c}{\frac{M}{2} - 1}, r+ \frac{d}{2}), & m > \frac{M}{2}
\end{cases}
\label{eq:points1}
\end{equation}
With $(point_1, . . . , point_N )$, we can generate composite queries using the following complete point formulation as in Eq.~\ref{eq:querypoints}: 

\begin{equation}
    Q^{(i)} = \varphi((point_1, . . . , point_M )^{(i)}) + (e_1, . . . , e_M )
\label{eq:querypoints}
\end{equation}

The $\varphi$ function in Eq~\ref{eq:querypoints} shows the dynamic point query update and is differentiable. This results in the best training convergence since each of the N control point content queries has its own explicit position prior.\\

\noindent
\textbf{The SAC2 Attention Block.} 
We use the Factorized Self- Attention (FSA) \cite{dong2021visual} in our model in accordance with~\cite{zhang2022text,ye2023dptext}.  FSA takes advantage of an intra-group self-attention ($SA_{intra}$) across M subqueries that correspond to each of the $Q(j)$ to capture the relationship between various points within each text instance. \textit{FastTextSpotter captures the relationship between various objects by introducing an inter-group self-attention ($SA_{inter}$) across K composite inquiries is used after $SA_{intra}$.} We hypothesize that the non-local self-attention mechanism, $SA_{intra}$, does not adequately capture the spherical shape of polygon control points. To address this, we incorporate local circular convolution~\cite{peng2020deep} to bolster factorized self-attention (FSA). Initially, $SA_{intra}$ processes to produce internal queries $Q_{intra} = SA_{intra}(Q)$, using identical keys as Q and values that exclude positional elements. Concurrently, locally enhanced queries are formed: $Q_{local} = ReLU(BN(CirConv(Q)))$. These are then integrated to create fused queries $Q_{fuse} = LN(FC(C + LN(Q_{intra} + Q_{local})))$, where C represents content queries acting as a shortcut, and $FC$, $BN$, and $LN$ denote fully connected layer, BatchNorm, and LayerNorm, respectively. Subsequently, $Q_{inter}$, which explores inter-positional relations, is derived from $Q_{fuse}$ using $SA_{inter}$ and passed to the deformable cross-attention module~\cite{zhu2020deformable}. Optimal performance and inference speed are achieved using the aforementioned training setup.
\begin{figure*}[t]
    \centering
    \includegraphics[width = \textwidth]{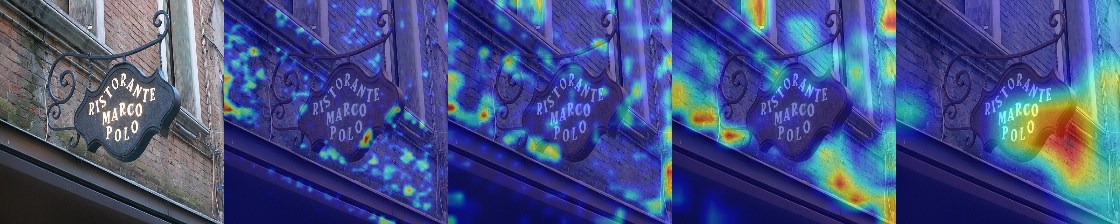}
    \caption{\textbf{Visualization of attention maps for Resnet-50 feature backbone}. (L) to (R) shows attention maps starting from the first layer.}
    \vspace{-0.302cm}
\end{figure*}
\begin{figure*}[t]
    \centering
    \includegraphics[width = \textwidth]{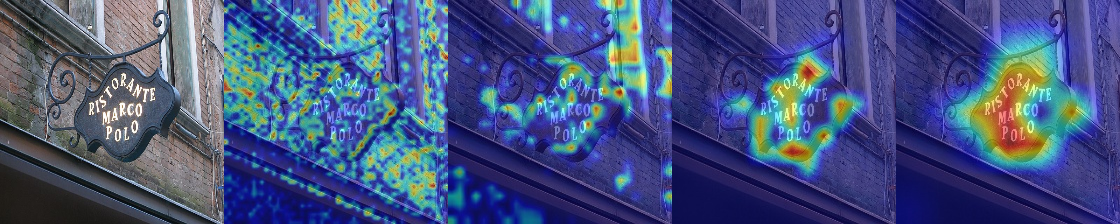}
    \caption{\textbf{Visualization of attention maps for Swin-Tiny feature backbone}. (L) to (R) shows attention maps from the first layer.}
    \label{fig:swin_feature_map}
\end{figure*}



\subsection{Loss Functions}
The overall losses used for FastTextSpotter can be summarised 
under the encoder $\mathcal{L}_{\text {enc }}$ and decoder $\mathcal{L}_{\text {dec }}$ blocks shown in eq.~\ref{eq:encoder} and eq.~\ref{eq:decoder} respectively.  


\begin{equation}
    \mathcal{L}_{\text {enc }}=\sum_j\left(\lambda_{\text {cls }} \mathcal{L}_{\text {cls }}^{(j)}+\lambda_{\text {coord }} \mathcal{L}_{\text {coord }}^{(j)}+\lambda_{\text {gloU }} \mathcal{L}_{\text {gloU }}^{(j)}\right)
\label{eq:encoder}
\end{equation}
\begin{equation}
    \mathcal{L}_{\text {dec }}=\sum_i\left(\lambda_{\text {cls }} \mathcal{L}_{\text {cls }}^{(i)}+\lambda_{\text {coord }} \mathcal{L}_{\text {coord }}^{(i)}+\lambda_{\text {char }} \mathcal{L}_{\text {char }}^{(i)}\right)
\label{eq:decoder}
\end{equation}

\noindent
Here, $\mathcal{L}{\text{cls}}^{(i)}$ represents the focal loss for text instance classification, while $\mathcal{L}{\text{coord}}^{(i)}$ denotes the L-1 loss used for control point coordinate regression. $\mathcal{L}{\text{char}}^{(i)}$ corresponds to the cross-entropy loss for character classification, and $\mathcal{L}{\text{gloU}}$ is the generalized IoU loss for bounding box regression, as defined in~\cite{rezatofighi2019generalized}. The weighting factors for these losses are represented by $\lambda_{\text{cls}}$, $\lambda_{\text{char}}$, $\lambda_{\text{coord}}$, $\lambda_{\text{cls}}$, and $\lambda_{\text{gloU}}$.

\noindent
\textbf{Instance classification loss.} We use the focal loss as the classification loss of text instances. For the i-th query, the loss is denoted as: 
\begin{equation}
\begin{split}
    \mathcal{L}_{\text {cls }} = -\mathds{1}_{\{i \in Pic(\sigma)\}} \alpha (1-\hat{ b}^{(i)})^{\gamma}\log(\hat{ b}^{(i)})) \\
    -\mathds{1}_{\{j \notin Pic(\sigma)\}} (1-\alpha) (\hat{ b}^{(j)})^{\gamma}\log(1-\hat{ b}^{(i)})
\end{split}
\label{eq:classification}
\end{equation}
\noindent
\textbf{Control Point Loss.} We used L-1 loss for control point regression the loss is defined for the i-th query:
\begin{equation}
     \mathcal{L}_{\text {coord }}^{(i)} = \mathds{1}_{\{i \in Pic(\sigma)\}} \sum_{j = 1}^{M}\left\|e_j^{(\sigma^{-1}(i))} - \hat{e}_j^{(i)}\right\|
\label{eq:controlp}
\end{equation}
\noindent
\textbf{Character classification Loss.} Character recognition seems like a classification problem, where each class is a specific character. We use cross-entropy loss, it defined as:
\begin{equation}
     \mathcal{L}_{\text {char }}^{(i)} = \mathds{1}_{\{i \in Pic(\sigma)\}} \sum_{j = 1}^{A}(-f_j^{(\sigma^{-1}(i))}\log{\hat{f}}_j^{(i)})
\label{eq:character}
\end{equation}

\section{Experimentations}
\subsection{Datasets}\label{dataset}
\begin{table*}[t]
\centering
\caption{ Results of scene text spotting on Total-Text and CTW1500. ''None '' denotes recognition without a lexicon. The ``Full'' lexicon contains all the words in the test set. Results style: \textbf{best}, \underline{second best}.}
\begin{tabular}{@{}cccccccccccc@{}}
\toprule
 & \multicolumn{5}{c}{Total-Text}&  \multicolumn{5}{c}{CTW1500} &\\ \cmidrule(lr){2-6} \cmidrule(lr){7-11}
Methods& \multicolumn{3}{c}{Detection}&  \multicolumn{2}{c}{End-to-end} & \multicolumn{3}{c}{Detection}&  \multicolumn{2}{c}{End-to-end} & FPS\\ \cmidrule(lr){2-4} \cmidrule(lr){5-6}\cmidrule(lr){7-9}\cmidrule(lr){10-11}
 & P & R& F& None& Full&  P & R & F& None& Full &\\ \hline
 Text Perceptron\cite{qiao2020text}&88.8 &81.8 &85.2&69.7&78.3 & 87.5& 81.9& 84.6& 57.0& -& -\\
ABCNet\cite{liu2020abcnet}&-& -& -& 64.2& 75.7 & -& -& 81.4&45.2& 74.1& 6.9 \\
MANGO\cite{qiao2021mango}&-&-&-&72.9&83.6&  - & -& -& 58.9 & 78.7&4.3\\
ABCNet v2\cite{liu2021abcnet} &90.2&84.1 &87.0&70.4 &78.1& 85.6& \underline{83.8}& 84.7&57.5& 77.2& 10 \\
Swintextspotter\cite{huang2022swintextspotter}& -& -& \textbf{88.0}&  74.3 & 84.1& -&  -&  \underline{88.0}&51.8& 77.0&-\\ 
Abinet++\cite{fang2022abinet++}&-&-&-&\underline{79.4} & 85.4& -& -& -& \textbf{61.5} &\underline{81.2}& 10.6 \\
TESTR\cite{zhang2022text}& \textbf{93.4}& 81.4& 86.90&73.25& 83.3& \underline{89.7}& 83.1& 86.3&53.3& 79.9&\underline{5.5}\\
DeepSolo~\cite{ye2023deepsolo} & \underline{93.1} & \underline{82.1} & \underline{87.3} & \textbf{79.7} & \textbf{87.0} &  & &  & \underline{60.01} & 78.4 & 10 \\ \hline
\textbf{FastTextSpotter}& 90.58& {\textbf{85.46}}& \underline{87.95} & 75.14&\underline{86.0}&{\textbf{91.45} }& \textbf{85.16} & \textbf{88.19} &56.02  &\textbf{82.91} &\textbf{5.38} \\
\bottomrule
\end{tabular}
\label{tab:my-table4}
\end{table*}
\noindent
For comparison with state-of-the-art methods, we selected the following benchmarks for experimental validation: \textbf{ICDAR 2015}\cite{karatzas2015icdar}, the official dataset of the ICDAR 2015 robust reading competition, is used for evaluating regular text spotting, employing the same test-train split as in the competition. \textbf{Total-Text}\cite{ch2020total} is a widely recognized benchmark for arbitrary-shaped text spotting, providing word-level text instances for evaluation. \textbf{CTW1500}\cite{liu2019curved} serves as another benchmark for arbitrary-shaped text spotting, featuring sentence-level text instances. \textbf{Vin-Text}\cite{nguyen2021dictionary} is a Vietnamese text dataset utilized for assessing the performance of multilingual text spotting systems.

 
\subsection{Implementation Details}
The hyper-parameters for the deformable transformer~\cite{zhu2020deformable} were configured similarly to the original implementation, with 8 attention heads and 4 sampling points utilized for the deformable attention mechanism. The number of layers for both the encoder and decoder was set to 6

\noindent
\textbf{Data augmentation.} During pre-training, we apply data augmentation by randomly resizing the images, with the shorter edge ranging from 480 to 896 pixels, while constraining the longest edge to a maximum of 1600 pixels. Additionally, an instance-aware random cropping technique is employed.

\noindent
\textbf{Pre-training.} We pre-trained FastTextSpotter using the SynthText150k~\cite{liu2020abcnet}, MLT 2017~\cite{nayef2017icdar2017}, and TotalText~\cite{ch2020total} datasets over 4000K iterations, starting with a learning rate of $2 \times 10^{-5}$, which was reduced by a factor of 0.1 after 3000K iterations. The AdamW~\cite{loshchilov2017decoupled} optimizer was employed, with parameters $\beta_1 = 0.9$, $\beta_2= 0.999$, and a weight decay of $10^{-5}$. We utilized $Q = 100$ composite queries, with 20 control points, and set the maximum text length to 25. The entire pre-training process was conducted on an RTX 3080 Ti GPU with a batch size of 1, spanning a total of 16 days.

\noindent
\textbf{Fine-tuning.} Following pre-training, we fine-tuned on the Total-Text and ICDAR2015 datasets for 200K iterations. For the CTW1500 dataset, FastTextSpotter was fine-tuned over 2000K iterations, with the maximum text length set to 100, given its sentence-level annotations that require additional training iterations. For the VinText dataset, the model was trained for 1000K iterations.

\subsection{Comparison with State-of-the-art}

\noindent

\begin{figure*}[t]
\centering
\begin{adjustbox}{max width=\textwidth}
\begin{tabular}{c c c c c c c c} 
\includegraphics[height=2cm, width= 2cm]{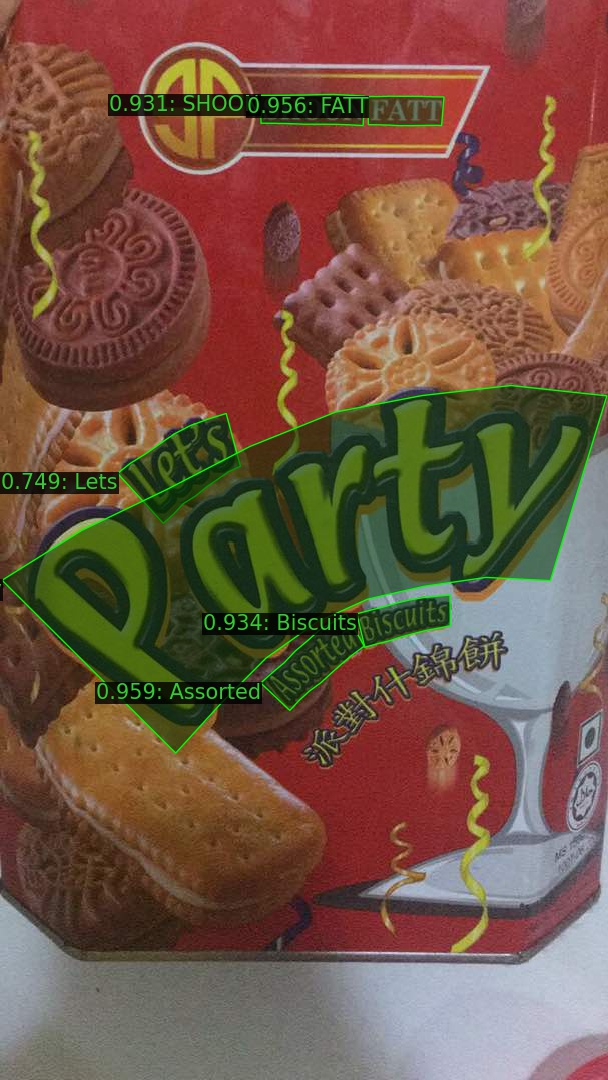}&
\includegraphics[height=2cm, width= 2cm]{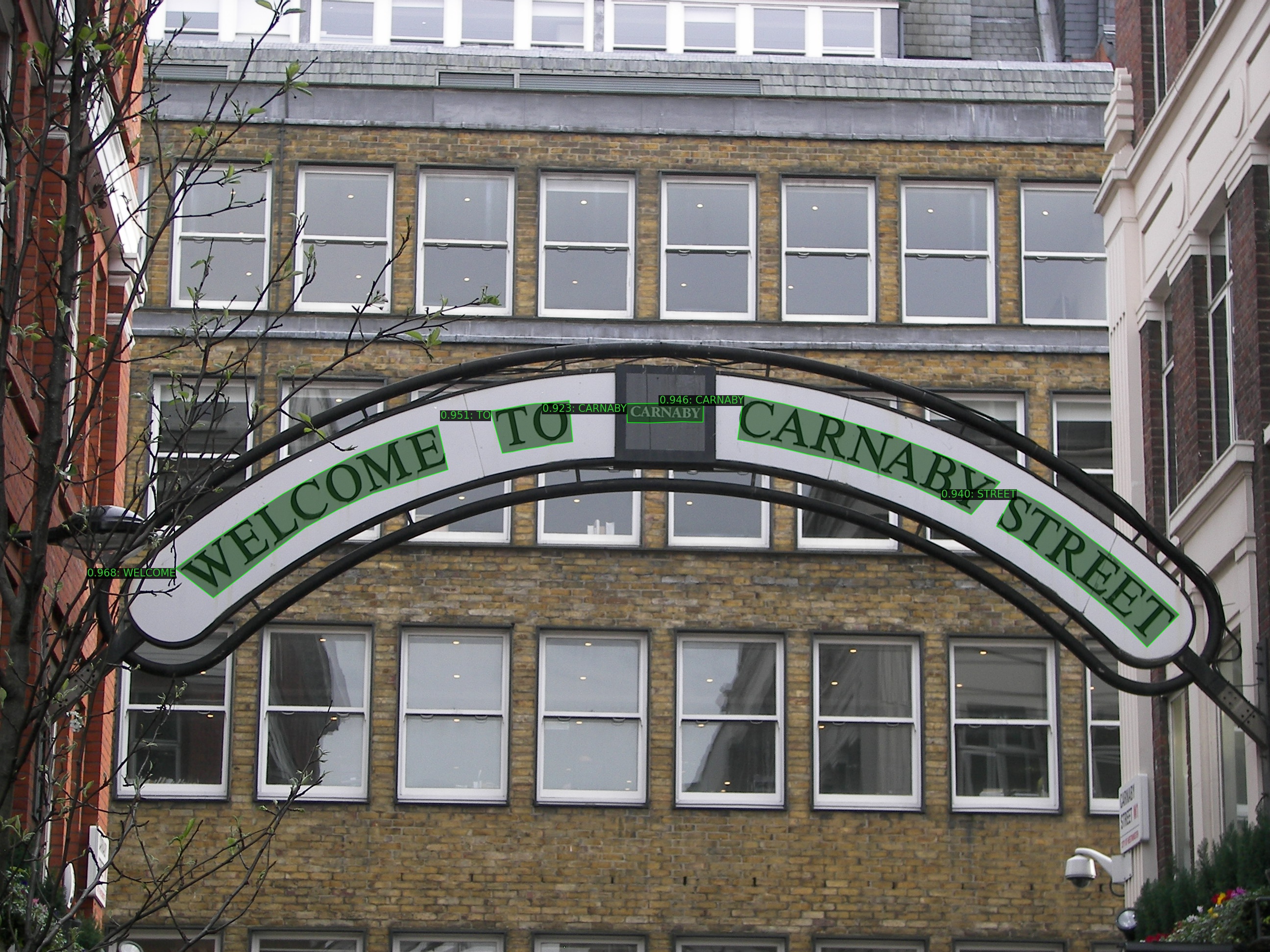}&
\includegraphics[height=2cm, width= 2cm]{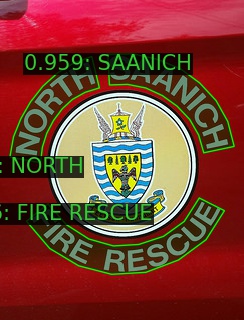}&
\includegraphics[height=2cm, width= 2cm]{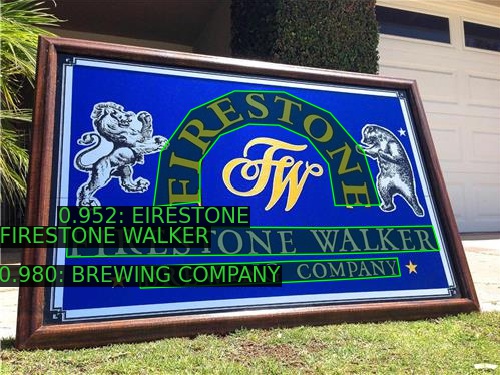}&
\includegraphics[height=2cm, width= 2cm]{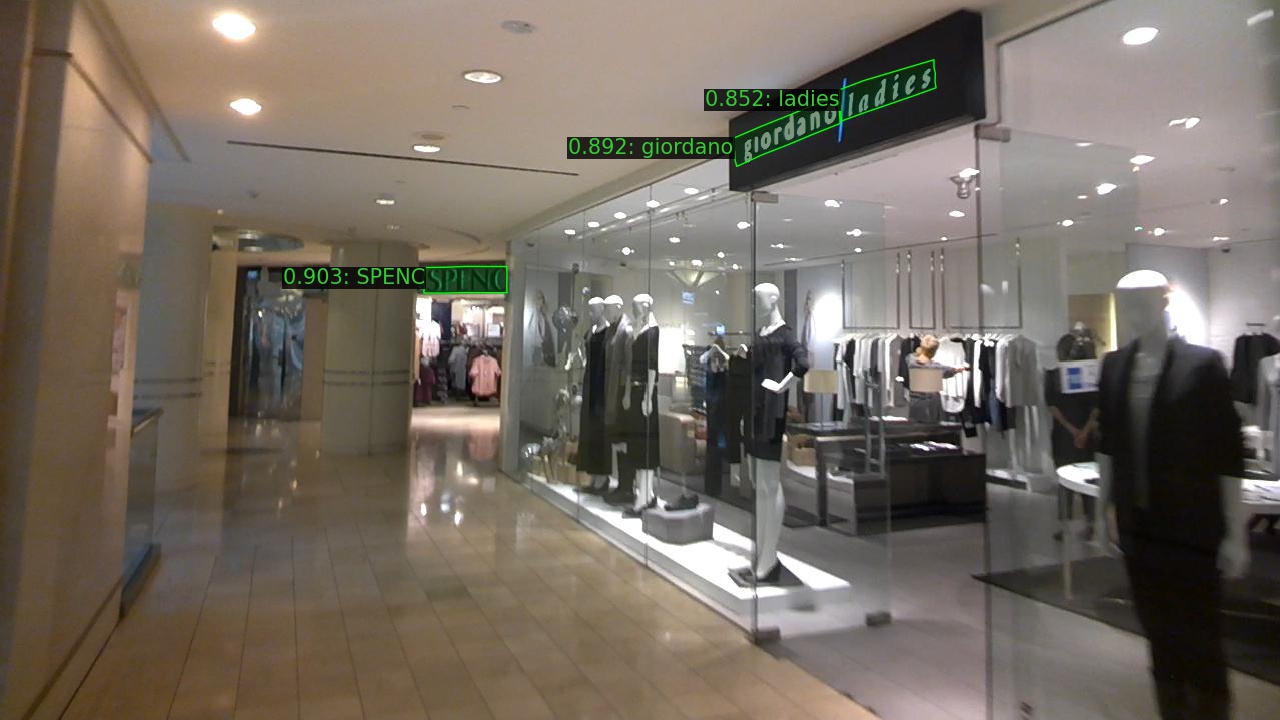}&
\includegraphics[height=2cm, width= 2cm]{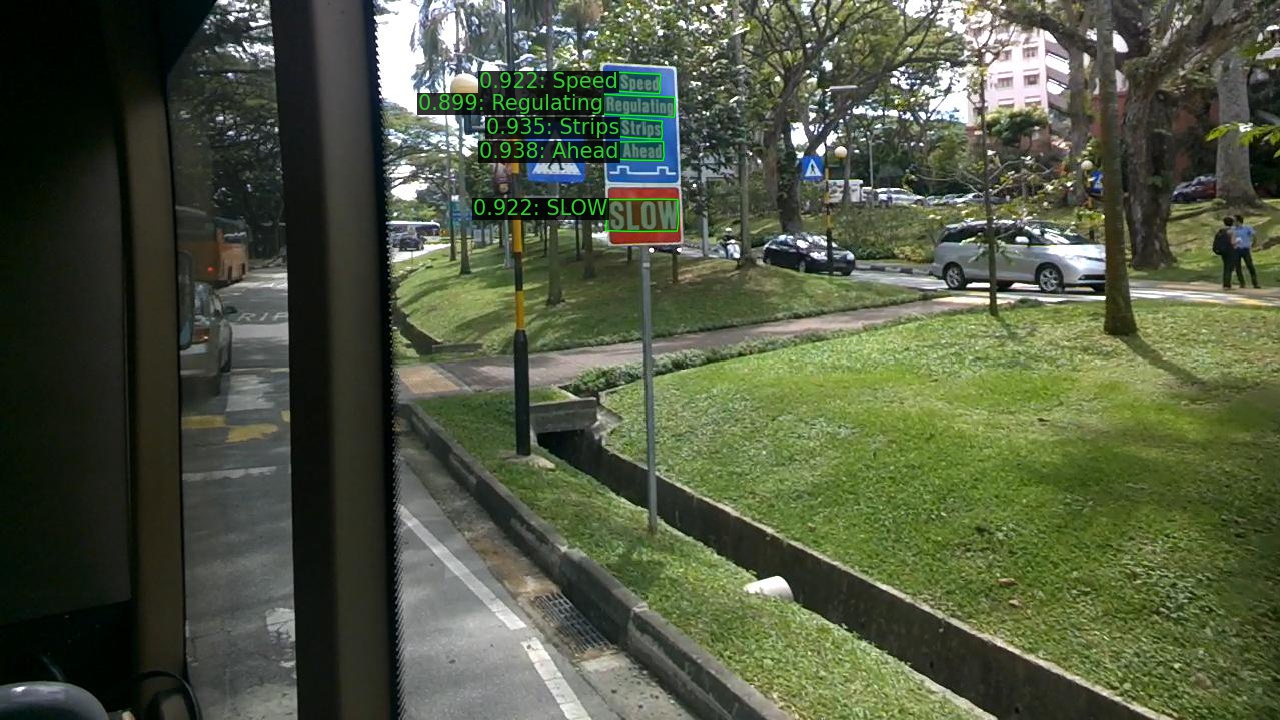}&
\includegraphics[height=2cm, width= 2cm]{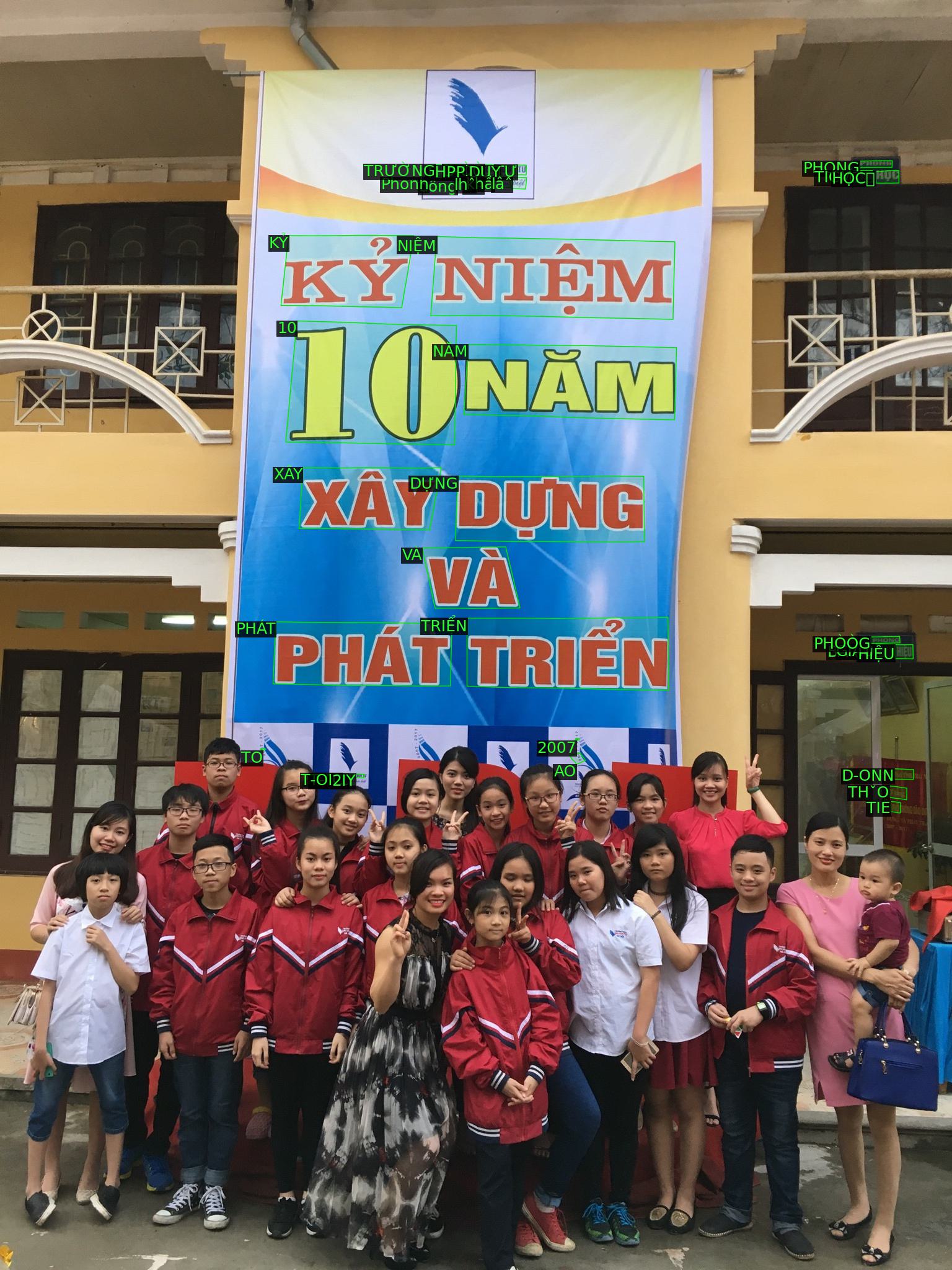}&
\includegraphics[height=2cm, width= 2cm]{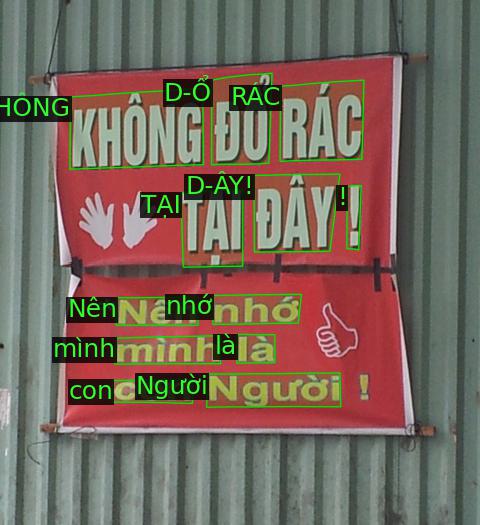}\\
\end{tabular}
\end{adjustbox}
\caption{Some illustration of our method on different datasets. Zoom in for better visualization. First two images from Total-Text, third and fourth images from CTW1500, fifth and sixth images from ICDAR15, and the last two images from Vintext.}
\label{fig:fig7}
\end{figure*}

Table~\ref{tab:my-table4} summarizes the results of FastTextSpotter compared to other text spotting methods. We surpass the state-of-the-art Abinet++\cite{fang2022abinet++} in end-to-end recognition on both the word-level Total-Text\cite{ch2020total} and sentence-level CTW1500~\cite{liu2019curved} benchmarks. For scene text detection, our model outperforms SwinTextSpotter~\cite{huang2022swintextspotter} on CTW1500 and achieves the second-best F-score on Total-Text. Notably, our model excels in recall metrics across both benchmarks. Qualitative examples are provided in Fig.~\ref{fig:fig7}.

\begin{table}[ht]
\centering
\caption{Results of scene text spotting ICDAR-15 datasets. “S”, “W”, and “G” are “Strong”, “Weak”, “Generic”, lexicas to recognise, respectively. Results style: \textbf{best}, \underline{second best}. }
\begin{adjustbox}{max width=\textwidth}
\begin{tabular}{ccccccc}
\toprule
\multicolumn{1}{c}{\multirow{2}{*}{Methods}}&\multicolumn{3}{c}{Detection}&\multicolumn{3}{c}{End-to-end}\\ \cmidrule(lr){2-4} \cmidrule{5-7}
& P& R & F&S& W& G\\ \hline

Text   Perceptron\cite{qiao2020text} & 89.4& 82.5& 87.1&80.5& 76.6& 65.1  \\ 
Unconstrained\cite{qin2019towards}& 89.4& 87.5& 87.5 & 83.4& 79.9& 68.0\\
MANGO \cite{qiao2021mango}& -& -& -&81.8& 78.9& 67.3  \\
ABCNet   v2 \cite{liu2021abcnet}& 90.4& 86.0& 88.1& 82.7& 78.5& 73.0  \\
Swintextspotter\cite{huang2022swintextspotter} & -& -& -&83.9& 77.3& 70.5  \\
TESTR\cite{zhang2022text}& 90.3&\textbf{89.7}& 90.0& 85.2& 79.4& 73.6  \\
Abinet++\cite{fang2022abinet++}& -& -& -&86.1& \textbf{81.9}& \textbf{77.8 } \\ 
PGNet\cite{wang2021pgnet} & 91.8& 84.8& 88.2& 83.3& 78.3& 63.5  \\
DeepSolo~\cite{ye2023deepsolo} & \underline{92.8} & \underline{87.4} &  \underline{90.10} &\textbf{86.8} & \bf{81.9} & \underline{76.9} \\ \hline
FastTextSpotter& \textbf{95.03} & 85.70  & {\textbf{90.13}} &\underline{86.63} & \underline{81.67} & 75.44 \\
\bottomrule
\end{tabular}
\end{adjustbox}
\label{tab:my-table5}
\end{table}

\noindent
We evaluated our method on the ICDAR15 benchmark~\cite{karatzas2015icdar} and compared it with state-of-the-art approaches (Table~\ref{tab:my-table5}). For text detection, we achieved a 5\% precision gain over TESTR~\cite{zhang2022text} and slightly exceeded the best F-measure. In text spotting, our method delivered the \textit{top performance in the challenging ``Strong'' category}, where each image contains a lexicon of only 100 words. We outperformed TESTR, SwinTextSpotter, and Abinet++ by roughly 1.5\%, 2.5\%, and 0.5\%. Figure~\ref{fig:fig7} shows our model's performance on this dataset in the third column.

\noindent
\textbf{Text Spotting in Low-Resourse.}We also evaluated Vintext~\cite{nguyen2021dictionary}, a low-resource benchmark for Vietnamese scene text detection, to demonstrate our model's generalizability. As shown in Table~\ref{tab:my-table6}, our method outperforms the state-of-the-art by nearly 2\%. Figure~\ref{fig:fig7} illustrates our model's performance on Vintext in the last column.\\

\noindent
\textbf{Why ABINet++ has better performance in Text Recognition?} ABINet++\cite{fang2022abinet++} and MANGO\cite{qiao2021mango} leverage linguistic information for text spotting, enhancing their performance in this metric. Despite relying solely on a visual approach, our method outperforms both.\\

\noindent
\textbf{Performance vs Efficiency Trade-off.} \textcolor{black}{Compared to previous approaches, our method shows optimum performance in terms of FPS which is \textbf{5.38} reported in Table~\ref{tab:my-table4}. Our FPS is almost halved in comparison to ABINet++ and DeepSolo~\cite{ye2023deepsolo} approach. \textcolor{black}{As depicted in Fig.~\ref{fig:fig1},  we observe the effectiveness of the SAC2 attention module introduced in the model (as shown in orange curve) when compared to the one with normal deformable attention~\cite{zhu2020deformable}. The key explanation behind this phenomenon is the usage of cyclic convolutions which was previously proposed for real-time instance segmentation~\cite{peng2020deep}.} Using this layer on top of the self-attention module along with the reference point resampling strategy used for both position and character queries helps in faster convergence during training and a better end-to-end spotting h-mean.}\\

\noindent
\textbf{Efficiency comparison with MANGO.} The MANGO text spotter~\cite{qiao2021mango} adopts a position-aware mask attention module to generate attention weights on each text instance and its characters to recognize character sequences without RoI operation. They achieve a slightly better FPS of 4.3 compared to ours owing to the fact that it's a single stage model with no self-attention. However, Table~\ref{tab:my-table4} and Table~\ref{tab:my-table5} highlight the fact that utilizing self-attention and transformer frameworks helps in significant improvement in metrics for detection and recognition tasks.

\begin{table}[ht]
\centering
\caption{Text recognition performance of the proposed and the state-of-the-art systems on the Vintext datasets. Results style: \textbf{best}, \underline{second best}.}
\begin{tabular}{cc}
\toprule
Methods& H-mean \\ \hline
ABCNet $+$ D~\cite{nguyen2021dictionary}& 57.4\\
ABCNet~\cite{liu2020abcnet}& 54.2\\
Mask Textspotter v3 $+$ D~\cite{nguyen2021dictionary}&68.5 \\
Swintextspotter\cite{huang2022swintextspotter}&\underline{71.1}\\
Mask Textspotter v3~\cite{nguyen2021dictionary}& 53.4 \\ \hline
FastTextSpotter(w/o fine-tune)&21.54\\
FastTextSpotter&\textbf{72.95}\\
\bottomrule
\end{tabular}

\label{tab:my-table6}
\end{table}

\subsection{Ablation Studies}
\noindent
We conducted ablation studies to assess the significance of the various components within the FastTextSpotter framework, leading to the following insights.\\

\noindent
\textbf{Swin Transformer serves as a robust visual backbone for text spotting.} 
We show a comparison of attention maps visualized in different layers for ResNet-50 backbone~\cite{he2016deep} when compared to the Swin-Tiny~\cite{liu2021swin} variant. The Swin attention better captures the global interactions between the different objects to localize the text region which helps them to get a substantial gain over ResNet-50 which primarily captures more local spatial relationships with the attention. The hierarchical shifted-window mechanism is highly useful to control the attention on top of the text region boundaries to have a more complete understanding using better contextual cues. More empirical results on the utility of the Swin backbone for the text spotting has been illustrated in previous works~\cite{das2024diving,das2024harnessing}. \\ 

\noindent
\textbf{Effect of Reference Point Resampling and SAC2 Module.}
\textcolor{black}{The effect of adding the SAC2 and reference point resampling strategies is shown in Table~\ref{tab:table3} where we clearly observe substantial gain in end-to-end recognition performance along with an optimal gain in detection too. Also using these modules helps us to gain faster convergence during the pre-training of FastTextSpotter.}

\begin{table}[t]
\centering
\caption{\textbf{Effectiveness of SAC2 and Reference Point Resampling Modules.} Performance obtained by pre-training model under this setting. ''LD'' and ''CD'' stand for Text Location Decoder and Text Recognition Decoder, respectively. 'None' indicates that no lexicon was used.}
\begin{adjustbox}{max width=\textwidth}
\begin{tabular}{@{}ccccccc@{}}
\toprule
\multicolumn{3}{c}{Modules}&\multicolumn{3}{c}{Detection}&End-to-End\\ \cmidrule(lr){1-3} \cmidrule(lr){4-6} \cmidrule(lr){7-7}
\multirow{2}{*}{\shortstack{Refference Points\\ Reampling} } & \multirow{2}{*}{\shortstack{SAC2\\in LD}} & \multirow{2}{*}{\shortstack{SAC2\\in CD}} &  \multirow{2}{*}{P} & \multirow{2}{*}{R} & \multirow{2}{*}{F} &\multirow{2}{*}{None}\\
& & & & &\\\hline
\multicolumn{1}{c}{\color{red}\xmark}&\multicolumn{1}{c}{\color{red}\xmark}&\multicolumn{1}{c}{\color{red}\xmark}&89.63&70.34&78.82&60.56\\
\multicolumn{1}{c}{\color{teal}\cmark}&\multicolumn{1}{c}{\color{red}\xmark}&\multicolumn{1}{c}{\color{red}\xmark}&91.36&72.52&80.85&62.38\\
\multicolumn{1}{c}{\color{teal}\cmark}&\multicolumn{1}{c}{\color{red}\xmark}& \multicolumn{1}{c}{\color{teal}\cmark}&91.75&74.35&82.13&65.46\\
\multicolumn{1}{c}{\color{teal}\cmark}&\multicolumn{1}{c}{\color{teal}\cmark}&\multicolumn{1}{c}{\color{red}\xmark}&\textbf{93.59} &75.20 &83.40 &67.06\\
\multicolumn{1}{c}{\color{teal}\cmark}& \multicolumn{1}{c}{\color{teal}\cmark}& \multicolumn{1}{c}{\color{teal}\cmark}&91.3&\textbf{77.33}&\textbf{83.68}&\textbf{68.87}\\
\bottomrule
\end{tabular}
\end{adjustbox}

\label{tab:table3}

\end{table}



\section{Conclusion and Future Work}

We proposed a novel efficient transformer model for text spotting,  FastTextSpotter, which not only establishes itself as a robust and efficient solution in the field of text spotting but also excels in operational efficiency. It outperforms previous state-of-the-art models in both end-to-end text recognition and scene text detection tasks, notably achieving top recall metrics for the Total-Text and CTW1500 benchmarks. The model’s efficiency is highlighted by its enhanced processing speed and reduced computational demands compared to the existing SOTA models, making it well-suited for real-time applications. Moreover, we show the effectiveness of the model for spotting in multiple languages, namely English and Vietnamese. Expanding its capabilities to include a broader array of languages, especially those with complex scripts could significantly increase its applicability and utility. \textcolor{black}{We plan to extend our evaluation to include diverse languages and scripts such as Arabic, Chinese, and Hindi, to further validate and enhance the model’s versatility and effectiveness in various linguistic contexts.}

\section*{Acknowledgements}
This work acknowledges the financial support of Department of Research and Universities of the Generalitat of Catalonia to the DocAI Research Group: Group on Document Intelligence (2021 SGR 01559), Grant PID2021-126808OB-I00 funded by MCIN/AEI/ 10.13039/501100011033 and by ERDF/EU and Ph.D. Scholarship from AGAUR (2023 FI-3-00223).
\bibliographystyle{splncs04}
\bibliography{egbib}
%




\end{document}